%% file: main.tex
\documentclass[letterpaper]{article} % DO NOT CHANGE THIS
\usepackage{aaai2026}  % DO NOT CHANGE THIS
\usepackage{times}  % DO NOT CHANGE THIS
\usepackage{helvet}  % DO NOT CHANGE THIS
\usepackage{courier}  % DO NOT CHANGE THIS
\usepackage[hyphens]{url}  % DO NOT CHANGE THIS
\usepackage{graphicx} % DO NOT CHANGE THIS
\usepackage{natbib}  % DO NOT CHANGE THIS AND DO NOT ADD ANY OPTIONS TO IT
\usepackage{caption} % DO NOT CHANGE THIS AND DO NOT ADD ANY OPTIONS TO IT
\usepackage{algorithm}
\usepackage{algorithmic}

\usepackage{newfloat}
\usepackage{listings}
\DeclareCaptionStyle{ruled}{labelfont=normalfont,labelsep=colon,strut=off} % DO NOT CHANGE THIS
\floatstyle{ruled}
\newfloat{listing}{tb}{lst}{}
\floatname{listing}{Listing}
\usepackage[utf8]{inputenc} % allow utf-8 input
\usepackage[T1]{fontenc}    % use 8-bit T1 fonts
\usepackage{url}            % simple URL typesetting
\usepackage{booktabs}       % professional-quality tables
\usepackage{amsfonts}       % blackboard math symbols
\usepackage{nicefrac}       % compact symbols for 1/2, etc.
\usepackage{microtype}      % microtypography
\usepackage{colortbl}

\usepackage{graphicx}
\usepackage{multirow}
\usepackage{amsmath}
\usepackage{caption}
\usepackage{algorithm}
\usepackage{algorithmic}
\usepackage{graphicx}
\usepackage{todonotes}
\usepackage{makecell}
\usepackage{xcolor} 
\usepackage{tikz} % 在导言区加上

\definecolor{HighlightColor}{gray}{0.9}
\definecolor{mygray}{gray}{.9}
\definecolor{ggray}{RGB}{127,127,127}
\definecolor{reda}{RGB}{192,0,0}
\definecolor{redb}{RGB}{217,148,143}
\definecolor{myyellow}{RGB}{190,144,0}
\definecolor{mygreen}{RGB}{80,100,40}
\definecolor{myblue}{RGB}{30,90,100}
\definecolor{mygray1}{RGB}{245,245,245}

\title{\textbf{InfinityHuman: Towards Long-Term \\Audio-Driven Human Animation}} 
\author{     % Authors     
    Xiaodi Li$^*$\textsuperscript{\rm 1},      
    Pan Xie$^*$\textsuperscript{\rm 1},      
    Yi Ren$^*$\textsuperscript{\rm 2},      
    Qijun Gan$^*$\textsuperscript{\rm 1,2},      
    Chen Zhang\textsuperscript{\rm 2},      
    Fangyuan Kong\textsuperscript{\rm 1}, \\     
    Xiang Yin\textsuperscript{\rm 1}$^\dagger$,      
    Bingyue Peng\textsuperscript{\rm 1},     
    Zehuan Yuan\textsuperscript{\rm 1} 
} 
\affiliations{     % Affiliations     
    \textsuperscript{\rm 1}ByteDance \quad \textsuperscript{\rm 2}Zhejiang University\\     
    \small $^*$Equal Contribution \quad $^\dagger$Corresponding Author \\
    \normalsize Project Page: \url{https://infinityhuman.github.io/}
}

\usepackage{bibentry}
\begin{document}
\twocolumn[{
\renewcommand\twocolumn[1][]{#1}
\maketitle
\vspace{-4mm}
\begin{center}
    \captionsetup{type=figure}
    \vspace{-4mm}
    \includegraphics[width=0.9\textwidth]{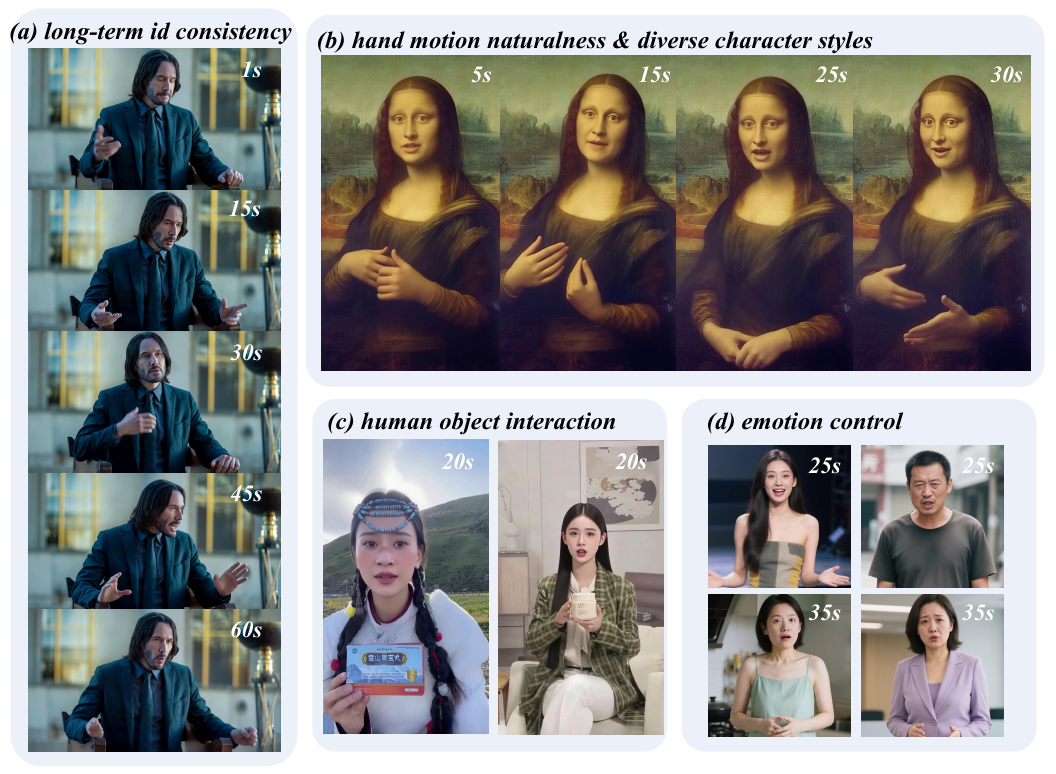}
    \vspace{-2mm}
    \caption{\textbf{InfinityHuman} is an audio-driven full-body animation framework that synthesizes long-duration videos with (a) temporally consistent visual appearance, (b) expressive and style-rich hand gestures, (c) dynamic human-object interactions, and (d) emotion-controllable, audio-aligned full-body motions.}
    \label{fig:teaser}
\end{center}
}]

\input{figures/fig2}

\input{sec/0_abstract}

\input{sec/1_intro}
\input{sec/2_related_work}

\input{sec/3_method}

\input{sec/4_experiment}

\input{sec/5_conclusion}

\bibliography{aaai2026}

\input{sec/7_supp}

\end{document}

%% file: figures/fig2.tex
\begin{figure*}[h]
    \centering
    \includegraphics[width=0.9\textwidth]{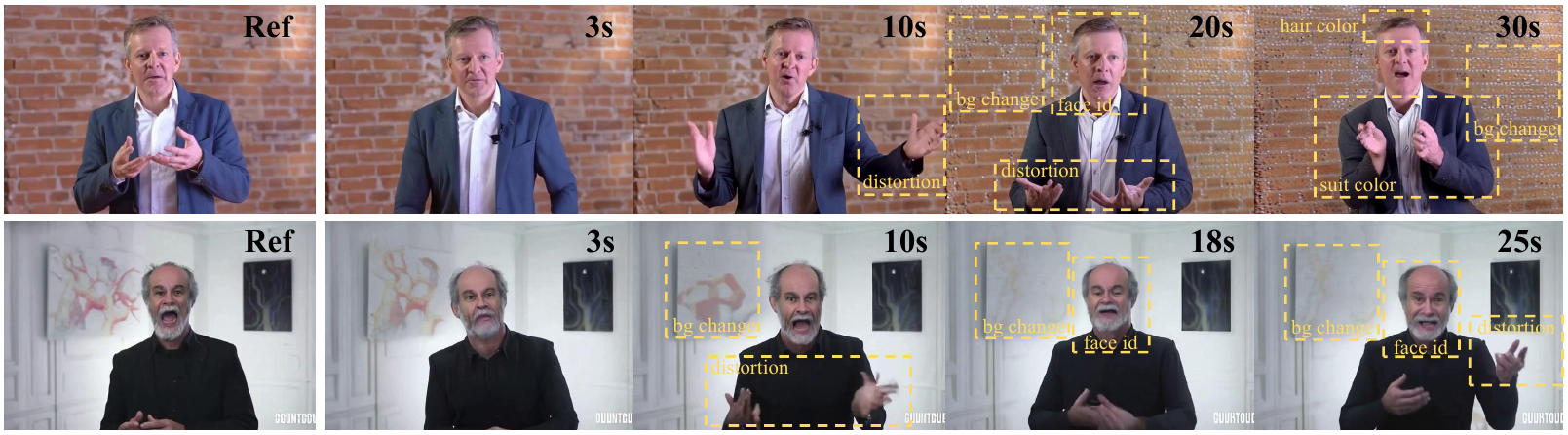}
    \vspace{-2mm}
\caption{\textbf{Progressive Degradation in Long Video Animation by Previous Methods.} Existing methods suffer from cumulative errors leading to pronounced identity drift (facial inconsistencies), color shifts (hair, clothing), scene instability (background fluctuations), and hand motion artifacts. These challenges underscore the necessity of InfinityHuman’s pose-guided refiner and hand-specific optimization for producing high-fidelity, temporally coherent animations over extended sequences.}

    \label{fig:fig2}
    \vspace{-4mm}
\end{figure*}

%% file: sec/0_abstract.tex
\begin{abstract}
% \vspace{-10pt}

Audio-driven human animation has attracted wide attention thanks to its practical applications. However, critical challenges remain in generating high-resolution, long-duration videos with consistent appearance and natural hand motions. Existing methods extend videos using overlapping motion frames but suffer from error accumulation, leading to identity drift, color shifts, and scene instability. Additionally, hand movements are poorly modeled, resulting in noticeable distortions and misalignment with the audio. In this work, we propose InfinityHuman, a coarse-to-fine framework that first generates audio-synchronized representations, then progressively refines them into high-resolution, long-duration videos using a pose-guided refiner. Since pose sequences are decoupled from appearance and resist temporal degradation, our pose-guided refiner employs stable poses and the initial frame as a visual anchor to reduce drift and improve lip synchronization. Moreover, to enhance semantic accuracy and gesture realism, we introduce a hand-specific reward mechanism trained with high-quality hand motion data. Experiments on the EMTD and HDTF datasets show that InfinityHuman achieves state-of-the-art performance in video quality, identity preservation, hand accuracy, and lip-sync. Ablation studies further confirm the effectiveness of each module. Code will be made public.
\end{abstract}

%% file: sec/1_intro.tex
\section{Introduction}
Audio-driven character animation aims to generate realistic human videos from a single image and audio input, transforming static portraits into speaking characters. This technology holds significant potential across various industries, including advertising, vlogging, and film production. With the rapid advancement of video generation models, recent research~\cite{zhang2023sadtalker, wang2024vexpress, lin2024cyberhost, diffted, cui2024hallo3, lin2025omnihuman,wang2025fantasytalking} has progressed from driving facial and head movements to full-body animation, greatly enhancing the expressiveness and richness of generated content.

% Recent advances in full-body human animation have made significant progress, but generating high-resolution, long-duration, and natural videos remains a challenging problem. The primary difficulties are as follows: \textbf{i) Long-Term Identity Consistency:} Existing methods typically use overlapping motion frames to extend video sequences. However, as the video length increases, errors accumulate over time, making it difficult to maintain consistent identity. This results in overall video degradation, with facial features gradually drifting (e.g., changes in eye color, eyebrow shape, and facial proportions), and inconsistencies in identity-preserving characteristics (e.g., mismatched hairstyle or clothing). \textbf{ii) Hand Motion Naturalness}: Previous work has mainly focused on facial naturalness and overall body movement, without providing detailed handling of hand movements, which involve small-range, high-speed motions. Consequently, large hand gestures tend to introduce distortions and artifacts, and the misalignment between hand movements and audio further diminishes the expressiveness and realism of the generated video.

% % by xiepan
Despite notable progress in full-body human animation, critical challenges remain in generating high-resolution, long-duration, and naturally coherent videos. These challenges can be grouped into two main areas: \textbf{i) Long-Term Visual Consistency}: Existing methods~\cite{lin2025omnihuman,wang2025fantasytalking,chen2025hunyuanvideoavatar,cui2024hallo3,kong2025let,gan2025omniavatar} typically extend video sequences using overlapping motion frames. However, as sequence length increases, accumulated errors undermine visual coherence, resulting in progressive degradation. This degradation manifests in three key aspects: inconsistent character identity (e.g., variations in facial proportions or clothing); global color incoherence (e.g., erratic shifts in tone or brightness); and scene instability (e.g., shifting or disappearing background objects). \textbf{ii) Hand Motion Naturalness}: Prior work has predominantly focused on facial naturalness and coarse body movements, neglecting the nuanced handling of hand motions—small-range yet high-speed movements. Consequently, large hand gestures frequently lead to distortions or artifacts, and misalignment between hand movements and audio further diminishes the expressiveness and realism of generated videos.

To address the aforementioned limitations, we propose \textbf{InfinityHuman}, a novel coarse-to-fine generation framework. This framework first produces low-resolution motion frames synchronized with audio, and subsequently outputs high-resolution long-form videos via a dedicated Refiner. 

% Our method introduces innovations in two key aspects: First, we design a \textbf{pose-guided refiner} to mitigate visual drift caused by error accumulation in long sequences. Pose sequences remain highly stable and exhibit minimal temporal degradation during long-duration generation, making them a reliable source of anatomical and dynamic information. Considering this, we adopt pose as a conditioning signal and leverage the initial frame as a stable visual reference to enhance visual consistency across time. Furthermore, pose information provides strong anatomical constraints and preserves fine-grained dynamic features such as lip movements, offering a reliable prior for reconstructing critical regions. With pose guidance, the model achieves more accurate lip-syncing, effectively alleviates small-scale motion distortions commonly observed in diffusion-based super-resolution methods, and significantly reduces hand-related artifacts such as finger overlapping.

% Since pose sequences are inherently stable and resistant to temporal degradation, we use them as reliable conditioning signals. by xiepan

Our method introduces innovations in two key aspects. First, we design a \textbf{pose-guided refiner} to address visual drift in long-duration sequences. %Since pose sequences are structurally decoupled from visual appearance, they inherently resist temporal degradation in appearance-related features. Thus we use them as reliable conditioning signals. 
Given that pose sequences are structurally decoupled from visual appearance, they inherently resist temporal degradation in appearance-related features. Consequently, we use them as reliable conditioning signals.
In addition, during continuous continuation, we incorporate the initial frame as a visual anchor to further enhance temporal consistency. This combination offers both dynamic guidance for maintaining temporal coherence and a reference for visual fidelity. Furthermore, compared vanilly diffusion-based super-resolution, the pose signal provides strong anatomical structure and preserves fine-grained motion patterns. This enables more accurate lip-syncing while effectively reducing common artifacts such as motion distortions and finger overlap in diffusion-based super-resolution.

% Secondly, considering that the human visual system is highly sensitive to hand distortions such as incorrect finger count, unnatural joint movements, we adopt a \textbf{hand-specific reward feedback mechanism} and incorporate high-quality hand motion data during training to guide hand generation. The mechanism encourages the model to produce semantically accurate, temporally consistent, and detailed gestures, thereby enhancing character expressiveness and the realism of the video.
%by xiepan
Secondly, considering that the human visual system is highly sensitive to hand distortions such as incorrect finger count, unnatural joint movements, we adopt a \textbf{hand-specific reward feedback mechanism} and incorporate high-quality hand motion data during training to guide hand generation. The mechanism encourages the model to produce temporally consistent and correct gestures, thereby enhancing character expressiveness and the realism of the video.

We evaluate InfinityHuman on the EMTD~\cite{echomimicv2} and HDTF~\cite{zhang2021flow} datasets, covering long-duration upper-body and talking-head scenarios. Qualitative and quantitative results show it achieves SOTA performance in video quality, id preservation, hand accuracy, and lip-sync. Ablation studies further confirm the effectiveness of our proposed model. Our contributions are summarized as follows:
% \begin{itemize}
%     \item We propose a coarse-to-fine full-body animation framework with a pose-guided refiner that mitigates visual drift and enhances temporal consistency in long sequences.
%     \item We introduce a hand-specific reward mechanism and integrate high-quality motion data to improve hand motion naturalness and audio alignment.
%     \item We achieve state-of-the-art results on both EMTD and HDTF datasets, demonstrating superior video quality, identity stability, and gesture realism across diverse scenarios.
% \end{itemize}
\begin{itemize}
    \item We propose InfinityHuman, a coarse-to-fine generation framework specifically designed to address the challenges of visual realism and temporal consistency in long-duration audio-driven character animation.
    % \item We develop a pose-guided refiner that leverages stable pose sequences as conditioning to correct accumulated errors, enhance lip-sync precision, and reduce artifacts in extended video generation.
    % by xiepan
    % \item We propose a pose-guided refiner that enhances diffusion-based super-resolution through explicit integration of pose conditioning and first-frame anchoring. The refiner leverages pose information to preserve temporal lip synchronization accuracy while utilizing the initial frame as a stable visual reference, thereby mitigating error propagation and ensuring long-term temporal coherence during extended sequence generation.
    \item We develop a pose-guided refiner that leverages stable pose sequences and the initial frame as a visual anchor to correct accumulated errors, maintain lip-sync accuracy, and reduce artifacts in extended video sequences.
    \item To improve hand movement realism and expressiveness, we introduce a hand-specific reward feedback mechanism, integrated with high-quality hand motion data.
    \item Comprehensive experiments on EMTD and HDTF datasets demonstrate that InfinityHuman outperforms state-of-the-art methods across multiple metrics.
\end{itemize}

%% file: sec/2_related_work.tex
\section{Related work}

\noindent{\textbf{Long Video Generation}} 
Existing methods~\cite{streamt2v, vidu, genlvideo, panda70m} extend video diffusion models to longer durations by modifying objectives or architectures. Autoregressive pipelines and memory modules~\cite{streamt2v, vidu} improve cross-segment consistency but require costly retraining on curated long-video datasets.
In contrast, training-free extensions such as Gen-L-Video~\cite{genlvideo} and FreeNoise~\cite{freenoise} improve efficiency via sliding-window attention and noise rescheduling. However, they offer limited temporal modeling, often causing temporal drift and less coherent transitions between segments.
To balance quality and efficiency, recent works~\cite{causvid, framepack, magi-1} fine-tune short-video diffusion models with previous motion frames as conditions for autoregressive continuation. Despite their flexibility, these methods suffer from error accumulation at inference, leading to degraded fidelity and identity shifts.
We adopt a similar strategy but address its limitations with a coarse-to-fine two-stage framework. A low-resolution long video is first generated, followed by a pose-guided refiner that corrects artifacts and restores spatial-temporal consistency, yielding high-resolution, identity-consistent long videos.

\noindent{\textbf{Audio-driven character animation.}} 
Recent advancements in audio-driven character animation have significantly improved lip-syncing and facial expression modulation using latent diffusion models. Works such as SadTalker \cite{zhang2023sadtalker} and Hallo \cite{xu2024hallo} enhance audio-to-facial synchronization with 3D rendering and diffusion techniques, while V-Express \cite{wang2024vexpress} and EchoMimic \cite{echomimicv2} refine naturalness by integrating audio with facial landmarks and control signals. Loopy \cite{jiang2024loopy} and OmniHuman-1 \cite{lin2025omnihuman} ensure identity consistency and mitigate data scarcity through multimodal training. Recent works have extended to full-body animation, with DiffTED \cite{diffted} introducing a one-shot framework for synchronized talking head and gesture animations, and CyberHost \cite{lin2024cyberhost} enhancing video quality using identity-independent features and human priors. Despite these advancements, generating high-resolution, long-duration, and natural videos remains a significant challenge, particularly in maintaining long-term identity consistency and ensuring the naturalness of hand motions. However, our Infinity Human leverages a pose-guided refiner and hand correction strategies to address these issues.

%% file: sec/3_method.tex
\section{Methodology}
\label{sec:method}
\textbf{Overview.}
As shown in Figure~\ref{fig:pipeline}, InfinityHuman is a unified framework designed to generate long-duration, full-body talking high-resolution videos \(V_{\text{hr}}\) from a single reference image \(I_{\text{ref}}\), audio \(\mathbf{c}_{\text{audio}}\), and an optional text prompt (\(\mathbf{c}_{\text{text}}\)), ensuring visual consistency, precise lip-sync, and natural hand movements. The framework adopts a coarse-to-fine strategy, starting with \textbf{Low-Resolution Audio-to-Video}(\S3.1) to produce coarse motion in \(V_{\text{lr}}\), followed by \textbf{Pose-Guided Refiner}(\S3.2) to generate high-resolution video \(V_{\text{hr}}\) conditioned on \(V_{\text{lr}}\) and \(I_{\text{ref}}\). Additionally, \textbf{Hand Correction Strategies}(\S3.3) are introduced to enhance the realism and structural integrity of hand movements.

\subsection{Low-Resolution Audio-to-Video}
\label{subsec:lowres_gen}
\textbf{Training Objective.}
We adopt Flow Matching~\cite{flowmatching} to train the low-resolution audio-to-video generation (LR-A2V). This approach enables efficient simulation of continuous-time dynamics by learning to predict the data’s velocity field. The backbone of our method is a DiT~\cite{peebles2023scalable}, denoted as $f_\theta$, which takes a noisy latent representation as input for all frames $z^{\mathrm{lr}}$, along with conditioning information from multiple modalities: a reference image $I_{\mathrm{ref}}$, text condition $c_{\mathrm{text}}$, audio condition $c_{\mathrm{audio}}$, and a continuous time step $t \in [0, 1]$. The low-resolution latent video $z^{\mathrm{lr}} = \{ z_i^{\mathrm{lr}} \}_{i=0}^f \in \mathbb{R}^{(f+1) \times h \times w \times c}$ is produced by encoding the input video $V_{\mathrm{lr}}$ using a 3D VAE encoder.

To construct training samples, Gaussian noise $\epsilon_i \sim \mathcal{N}(0, I)$ is sampled independently for each latent, and the noisy latent at diffusion time $t$ for latent $i$ is obtained by the diffusion process: 
\begin{equation}
z_{i,t}^{\mathrm{lr}} = \phi(z_i^{\mathrm{lr}}, t) = (1 - t) \cdot \epsilon_i + t \cdot z_{i,1}^{\mathrm{lr}}
\end{equation}

The target velocity is then defined as: 
\begin{equation}
v_{i,t} = \frac{d z_{i,t}^{\mathrm{lr}}}{d t} = z_{i,1}^{\mathrm{lr}} - \epsilon_i
\end{equation}

The DiT model is trained to predict these velocities for all frames jointly. The training objective minimizes the expected squared error:
% \begin{equation}
% \mathcal{L} = \mathbb{E}_{\epsilon_i \sim \mathcal{N}(0, I),\, t \sim \mathcal{U}(0,1)}
% \left\| f_\theta\bigl(\{ z_{i,t}^{\mathrm{lr}} \}_{i=0}^f, I_{\mathrm{ref}}, c_{\mathrm{text}}, c_{\mathrm{audio}}, t \bigr) - \{ v_{i,t} \}_{i=0}^f \right\|_2^2.
% \end{equation}
\begin{align}
\mathcal{L} = \mathbb{E}_{\epsilon_i \sim \mathcal{N}(0, I),\, t \sim \mathcal{U}(0,1)} \ 
& \left\| f_\theta\bigl(\{ z_{i,t}^{\mathrm{lr}} \}_{i=0}^f,\, I_{\mathrm{ref}},\, 
c_{\mathrm{text}},\, c_{\mathrm{audio}},\, t \bigr) \right. \notag \\
& \left. - \{ v_{i,t} \}_{i=0}^f \right\|_2^2
\end{align}

 % This formulation enables the model to learn a temporally coherent and identity-consistent mapping from noise to video, conditioned on the reference image, text prompts, and audio inputs. 
 % To support long-range continuation, motion frames from the previous segment are concatenated as conditioning input.

\input{figures/pipeline}

\noindent\textbf{Multimodal Condition Attention.} To improve the incorporation and alignment of audio information, we decouple the audio condition from other modalities by introducing a separate cross-attention branch specifically for audio. Formally, the identity-aware cross-attention is extended as follows:
\begin{equation}
\text{CA}_{\mathrm{mm}}\big(x^{\mathrm{lr}}, c_{\text{text}}, c_{\text{audio}}\big) = \mathrm{CA}\big(x^{\mathrm{lr}},\, c_{\text{text}} \big) + \mathrm{CA}\big(x^{\mathrm{lr}},\, c_{\text{audio}}\big)
\end{equation}
In this way, we enable more precise control over multimodal interactions, allowing the model to better align audio cues with visual dynamics and enhance the generation quality.

\subsection{Pose-Guided Refiner}
\label{subsec:pose_guided_refiner}

In long-term generation tasks, low-resolution video \( V_{\text{lr}} \) tends to accumulate errors over time, resulting in visual drift where the appearance deviates from the reference image \( I_{\text{ref}} \). To address this issue, the Pose-Guided Refiner (PG-Refiner) leverages the reference image \( I_{\text{ref}} \) as an identity prior and conditions on the low-resolution video \( V_{\text{lr}} \) along with its corresponding pose sequence \( \mathcal{P} = \{ p_i \}_{i=0}^{4f+1} \). This ensures both temporal coherence in motion and consistent appearance throughout the whole video.

\noindent\textbf{Low-Resolution Video Latent Condition.}  
To simulate the temporal degradation phenomenon, we filter out high-frequency signals from the low-resolution latent representation (\( z^{\text{lr}} \)) using a low-pass filter (LPF), and introduce noise augmentation to improve the model's ability to recover details and correct structural errors. Specifically, the degraded latent representation \( z^{\text{deglr}} \) is computed as:
\begin{equation}
z^{\text{deglr}} = \text{LPF}(z^{\text{lr}}) + \alpha_{\text{deg}} \cdot \epsilon
\end{equation}
where \( \text{LPF}(z^{\text{lr}}) \) extracts the low-frequency components of the video latent, \( \epsilon \sim \mathcal{N}(0, \sigma^2) \) is additive Gaussian noise, and \( \alpha_{\text{deg}} \) controls the noise strength.

% This process simulates the temporal degradation seen during generation, enabling the model to better handle accumulated errors and maintain visual fidelity and identity consistency over long sequences.

\noindent\textbf{Pose Guidance Condition.}
Considering that pose sequence information possesses strong structural properties, preserves fine-grained motions such as lip movements, and remains highly stable with minimal error accumulation in long-duration generation tasks, we adopt it as the condition. 
% Specifically, the inherent strong anatomical constraints in the pose sequence enable effective correction of complex hand poses, preventing issues such as finger overlap or unnatural hand configurations. Moreover, degradation caused by low-resolution input particularly affects subtle motions like lip movements, making pose information crucial for maintaining detailed synchronization and temporal coherence.

Based on this, we extract human and background keypoints from \( V_{\text{lr}} \), forming a pose sequence \(\{ p_i \}_{i=0}^{4f+1} \). To avoid scale mismatch and keypoint overlap across different resolutions, we use an 8-channel pixel-level representation: the first 7 channels encode human keypoints, and the last channel encodes up to 20 background keypoints. The resulting pose tensor is denoted as \( \mathcal{P} \in \mathbb{R}^{(4f+1) \times 4h \times 4w \times 8} \). Accordingly, we apply patchification along the temporal and spatial dimensions: the temporal axis is divided into \( f + 1 \) segments, and the spatial dimensions into \( h \times w \) patches, yielding pose tokens \( \mathcal{P}' \in \mathbb{R}^{(f+1) \times h \times w \times (64 \times 8)} \).

These pose tokens are projected into the latent space via a learned projection and fused with the high-resolution latent feature \( z^{\mathrm{hr}} \), producing a pose-aware latent representation \( z'^{\mathrm{hr}} = z^{\mathrm{hr}} + \mathrm{Proj}(\mathcal{P}') \). The resulting \( z'^{\mathrm{hr}} \) serves as the input to the generator, enhancing both visual fidelity and the temporal consistency of motion in the generated video.

\input{table/baseline}
\noindent\textbf{Refiner.} To further enhance temporal consistency, we utilize the initial reference frame as a visual anchor. The Refiner module leverages the reference image \( I_{\mathrm{ref}} \), pose conditional features \( P \), and the low-resolution degraded latent feature \( z_{\mathrm{deglr}} \) to generate high-resolution video frames. Since the model is enhanced with temporal degradation during training and introduces pose information as a control signal that is more direct and structurally informative than audio, it can effectively maintain long-term identity consistency with the assistance of the reference image.

Unlike previous methods~\cite{zeng2024make,hu2023animate} that rely on structure-aligned reference networks, we adopt a prefix-latent reference strategy to ensure identity consistency and enable high-quality long-sequence continuation. This strategy fully exploits the 3D global attention mechanism in the DiT architecture, allowing the model to directly extract identity features from the prefix latent.
Specifically, we denote the high-resolution latent sequence as \( \{ z_i^{\mathrm{hr}} \}_{i=0}^{f} \), where \( z_0^{\mathrm{hr}} = E(I_{\mathrm{ref}}) \) is the prefix latent extracted from the reference image using a pretrained 3D VAE encoder \( E(\cdot) \), and \( z_1^{\mathrm{hr}} \) to \( z_m^{\mathrm{hr}} \) represent motion latents from preceding segments. As the first frame is not temporally compressed, \( z_0^{\mathrm{hr}} \) preserves more detailed information crucial for identity preservation.

During forward diffusion, we inject noise $\epsilon_i \sim \mathcal{N}(0, I)$ only into the future latents:

% \begin{equation}
% z_{i,t}^{\mathrm{hr}} =
% \begin{cases}
% z_i^{\mathrm{hr}}, & 0 \le i \le m,\\[6pt]
% (1 - t) \cdot \epsilon_i + t \cdot z_i^{\mathrm{hr}},\quad \epsilon_i \sim \mathcal{N}(0,I), & m < i \le f,
% \end{cases}
% \end{equation}

\begin{equation}
z_{i,t}^{\mathrm{hr}} =
\begin{cases}
z_i^{\mathrm{hr}}, & 0 \le i \le m, \\[6pt]
(1 - t) \cdot \epsilon_i + t \cdot z_i^{\mathrm{hr}}, & m < i \le f
\end{cases}
\end{equation}

\noindent so that frames \(0\) through \(m\) remain noise-free to provide stable identity and motion guidance, and their noise predictions are excluded from the loss to maintain reference stability and preserve identity consistency.

Formally, the training objective minimizes the velocity prediction error over the noised subset:
% \begin{equation}
% \begin{aligned}
% \mathcal{L}_{\text{ref}} =
% \mathbb{E}_{\epsilon_i \sim \mathcal{N}(0, I),\, t \sim \mathcal{U}(0,1)} \;
% \mathbf{w} \cdot 
% \Biggl\|
% f_\theta\bigl(\{ \mathbf{z}_{i,t}^{\mathrm{hr}} \}_{i=0}^f,\, \mathbf{z}^{\mathrm{deglr}},\, P',\, I_{\mathrm{ref}},\, t \bigr)
% -
% \bigl\{ \mathbf{z}_{i,1}^{\mathrm{hr}} - \boldsymbol{\epsilon}_i \bigr\}_{i=0}^f
% \Biggr\|_2^2,
% \end{aligned}
% \end{equation}
\begin{align}
\mathcal{L}_{\text{ref}} = 
\mathbb{E}_{\epsilon_i \sim \mathcal{N}(0, I),\, t \sim \mathcal{U}(0,1)} \,
& \mathbf{w} \cdot \Biggl\|
f_\theta\bigl(\{ \mathbf{z}_{i,t}^{\mathrm{hr}} \}_{i=0}^f,\, \mathbf{z}^{\mathrm{deglr}},\, P',\, I_{\mathrm{ref}},\, t \bigr) \notag \\
& - \bigl\{ \mathbf{z}_{i,1}^{\mathrm{hr}} - \boldsymbol{\epsilon}_i \bigr\}_{i=0}^f
\Biggr\|_2^2
\end{align}

where \(\mathbf{w} = \{w_i\}_{i=0}^f\) is a mask vector defined as

\begin{equation}
w_i = 
\begin{cases}
1, & i > m, \\
0, & \text{otherwise}
\end{cases}
\end{equation}
To achieve the continuity of motion in long video generation during inference, we take the first \( m \) latents of a new chunk from the last \( m \) latents of the previous chunk, ensuring smooth motion transitions between chunks.

\subsection{Hand-Specific Reward Feedback Learning}
\label{subsec:hand_correction}
\input{figures/baseline}
Previous models primarily focus on body and facial movements but overlook detailed hand modeling, leading to unnatural hand distortions in generated videos. To address this, we introduce a hand-specific correction strategy that explicitly targets these artifacts.

The human visual system is highly sensitive to hand structures, with clear perceptual boundaries for distortions such as incorrect finger count, unnatural articulation, or broken textures. Motivated by this, we introduce a preference fine-tuning strategy that directly targets hand realism. By optimizing the diffusion model using reward scores from a pretrained image-level evaluator, we significantly improve the structural fidelity and visual quality of generated hands.

% Specifically, we leverage a pretrained image-level reward model to assess hand realism. To adapt it for video, we decode the low-resolution latent sequence $\{ z_{m,1}^{\mathrm{lr}} \}_{m=0}^f$ into RGB frames and randomly select one frame $X_i^{\mathrm{lr}}$ for evaluation. The training objective becomes:
% by xiepan
Specifically, we first manually constructed a dataset of 10,000 paired image data of hand structures. Leveraging this carefully domain-specific curated dataset, we performed fine-tuning on the open-source MPS~\cite{zhang2024learning} model to enhance its initial capability in capturing hand structural characteristics. Building on this, we leverage the pretrained image-level reward model to assess hand realism. To adapt it for video, we decode the low-resolution latent sequence $\{ z_{m,1}^{\mathrm{lr}} \}_{m=0}^f$ into RGB frames and randomly select one frame $X_i^{\mathrm{lr}}$ for evaluation. The training objective becomes:
% \begin{equation}
% \mathcal{L}_{\mathrm{hand}}(\theta) = \mathbb{E}_{c \sim p(c)}\,\mathbb{E}_{X_i^{\mathrm{lr}} \sim \mathcal{D}(z_{i,1}^{\mathrm{lr}})} \left[ - r_{\mathrm{hand}}(X_i^{\mathrm{lr}}, c) \right]
% \end{equation}
% where $X_i^{\mathrm{lr}}$ is a decoded frame randomly sampled from the low-resolution latent trajectory, and $r_{\mathrm{hand}}(\cdot)$ denotes the pretrained reward model’s assessment of hand quality. This approach introduces fine-grained, hand-specific supervision without additional annotations, effectively enhancing anatomical plausibility and reducing common distortions in generated human videos.

\begin{equation}
\mathcal{L}_{\mathrm{hand}}(\theta) = \mathbb{E}_{c \sim p(c)}\,\mathbb{E}_{X_i^{\mathrm{lr}} \sim \mathcal{D}(z_{i,1}^{\mathrm{lr}})} \left[ T - r_{\mathrm{hand}}(X_i^{\mathrm{lr}}, c) \right]
\end{equation}
where $X_i^{\mathrm{lr}}$ is a decoded frame randomly sampled from the low-resolution latent trajectory, $r_{\mathrm{hand}}(\cdot)$ denotes the pretrained reward model’s assessment of hand quality, and T denotes the threshold for hand quality. This approach introduces fine-grained, hand-specific supervision without additional annotations, effectively enhancing anatomical plausibility and reducing common distortions in generated human videos.

%% file: figures/pipeline.tex
\begin{figure*}[t]
    \centering
    \includegraphics[width=0.9\textwidth]{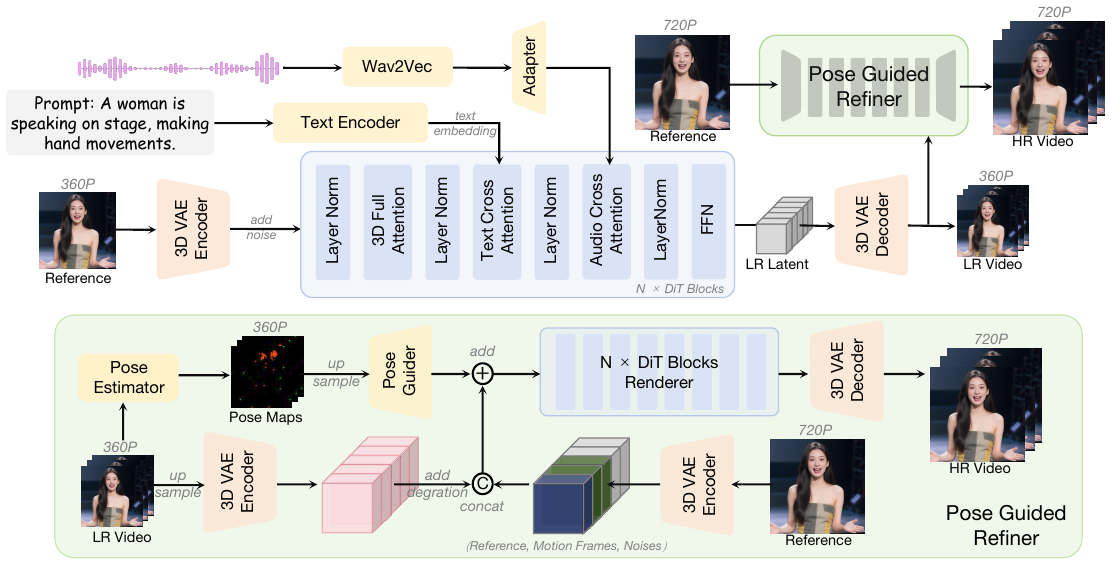}
    \vspace{-2mm}
    \caption{\textbf{InfinityHuman Pipeline.} The pipeline generates high-resolution (HR) audio-driven full-body videos through a two-stage coarse-to-fine process. First, a speech-aligned low-resolution (LR) video is generated using multimodal conditioning (text and audio) and DiT blocks. In the second stage, a pose-guided refiner utilizes pose guidance, LR latents, and reference images to restore degraded details, enhancing identity consistency, motion coherence, and hand realism.}
    \label{fig:pipeline}
    \vspace{-4mm}
\end{figure*}

%% file: table/baseline.tex
\begin{table*}[!h]
\vspace{-2mm}
\small
\centering
% \caption{\textbf{Quantitative Comparison of Audio-Driven Animation Methods on EMTD and HDTF}. \textbf{$*$} denotes methods limited to talking-head animation. InfinityHuman achieves SOTA results across benchmarks.(\S\ref{sec:sota_comparison})}
\vspace{-2mm}
% \label{tab:baseline}
\resizebox{\textwidth}{!}{
\begin{tabular}{|r|cccc|cc|c|cc|}
\Xhline{1.3pt}
\rowcolor{mygray}
 & \multicolumn{4}{c|}{{Video Quality}} & \multicolumn{2}{c|}{{Lip Sync}} & \multicolumn{1}{c|}{{ID}} & \multicolumn{2}{c|}{{Hand Stability}} \\
\rowcolor{mygray}
\multirow{-2}*{Method~~~~~~~} & FID$\downarrow$ & FVD$\downarrow$ & IQA$\uparrow$ & ASE$\uparrow$ & SYNC$\uparrow$ & SYND$\downarrow$ & FSIM$\uparrow$ & HKC$\uparrow$ & HKV \\
\hline\hline

\multirow{11}{*}{\rotatebox{90}{\makecell{\scriptsize ~~~~~~~~~~~~~~~~~~~~~~~~~~~~EMTD~~~~~~~~~~~~~~~~~~~~~~~~~ HTDF~~~ ~~~~~~~~~~~~~~~~~~~~~~~}}}

SadTalker$^*$~\cite{zhang2023sadtalker} & 147.73&	862.83&	1.72&	1.07&	\textbf{8.87}&	\textbf{6.71}&	\textbf{0.93} & - & - \\
AniPortrait$^*$~\cite{wei2024aniportrait} & 96.12 & 645.72 & 1.96 & 1.15 & 7.64 & 7.79 & 0.85 & - & - \\
V-Express$^*$~\cite{wang2024v} & 119.45 & 748.57 & 1.32 & 1.16 & 7.92 & 7.96 & 0.89 & - & - \\
EchoMimic$^*$~\cite{chen2025echomimic} & 167.17 & 757.38 & 1.61 & 1.19 & 6.71 & 8.23 & 0.82 & - & - \\
HyAva~\cite{chen2025hunyuanvideoavatar} & 100.10&	662.61&	1.52&	1.06&	7.22&	8.98	& 0.85& - & - \\
Hallo3~\cite{cui2024hallo3} & 74.10&	250.12&	1.95&	1.14&	7.31&	9.30&	0.91& - & - \\
MultiTalk~\cite{kong2025let} & 85.01&	404.45&	1.78&	1.13&	8.76&	7.69&	0.84& - & - \\
OmniAvatar~\cite{gan2025omniavatar} & 131.69&	705.14&	1.67	&1.10	&8.81	&7.76	&0.78 & - & -\\

\rowcolor{mygray1}
Ours & \textbf{69.28} & \textbf{239.05} & \textbf{2.11} & \textbf{1.22} & 8.59 & 7.53 & 0.89 & - & - \\
\hline\hline

Fantasy~\cite{wang2025fantasytalking}
& 133.73&	1307.20&	2.11&	1.12&	1.11&	12.88&	0.59&	0.57&	8.0\\
HyAva~\cite{chen2025hunyuanvideoavatar} & 139.39&	2160.92&	1.76&	1.18&	4.89&	9.37&	0.67&	0.75&	29.2 \\
Hallo3~\cite{cui2024hallo3} & 104.51&	1256.10&	2.31&	1.48&	4.26&	10.22&	0.73&	0.77&	6.3 \\
MultiTalk~\cite{kong2025let} & 103.68	&1040.43&	2.07&	1.30&	6.34&	8.47&	0.71&	0.79&	14.6 \\
OmniAvatar~\cite{gan2025omniavatar} & 82.54&	1104.99&	2.16&	1.31&	5.40&	9.13&	0.72&	0.86&	28.7 \\
\rowcolor{mygray1}
Ours & \textbf{60.71} & \textbf{979.88} & \textbf{2.48} & \textbf{1.59} & \textbf{6.56} & \textbf{7.97} & \textbf{0.84} & \textbf{0.90} & 16.0 \\

\hline
\end{tabular}
}
\caption{\textbf{Quantitative Comparison of Audio-Driven Animation Methods on EMTD and HDTF}. \textbf{$*$} denotes methods limited to talking-head animation. InfinityHuman achieves SOTA results across benchmarks.(\S\ref{sec:sota_comparison})}
\vspace{-4mm}
\label{tab:baseline}
\end{table*}

%% file: figures/baseline.tex
\begin{figure*}[h]
\vspace{-10pt}
    \centering
    \includegraphics[width=0.9\textwidth]{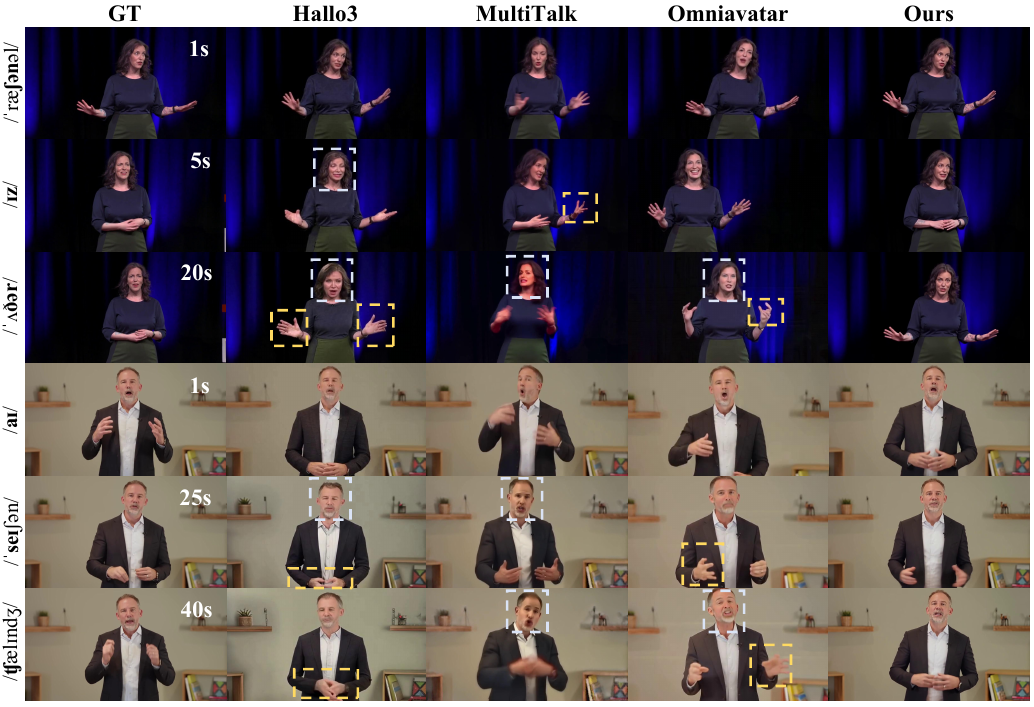}
\vspace{-2mm}
\caption{\textbf{Qualitative Results of Audio-Driven Animation Methods on EMTD.} Yellow and blue boxes highlight hand distortions and face ID mismatches, respectively. The results demonstrate the superiority of InfinityHuman in maintaining identity consistency, lip-sync accuracy, and visual fidelity during long-duration generation. Please zoom in for details. (\S\ref{sec:sota_comparison})}
\label{fig:baseline_sota}
\vspace{-4mm}
\end{figure*}

%% file: sec/4_experiment.tex
\section{Experiment}
\subsection{Implementation Details}
% by xiepan
\noindent\textbf{Datasets.} Our data processing pipeline is as follows: First, we employ SceneDetect~\cite{PySceneDetect} for temporal cropping of the raw videos. Next, we use YOLO~\cite{Jocher_Ultralytics_YOLO_2023} to track the single person, obtain corresponding spatiotemporal bounding boxes, and perform spatiotemporal cropping. Additionally, videos are filtered based on criteria including video quality, aesthetics, motion amplitude, hand clarity, mouth clarity, and the proportion of the person within the frame. Ultimately, this process yields 7,700 hours of single-person video clips, which is used to train the pose-guider refiner. Building on this dataset, SyncNet~\cite{chung2017out} is further employed to assess the synchronization between audio and mouth movements, filtering to obtain 1,800 hours of clips for training low-resolution audio-driven video generation, where each clip is 4 seconds.

% by xiepan
\noindent\textbf{Training.} To train audio-driven low-resolution video generation, we begin with a pretrained Goku-I2V~\cite{chen2025goku} model. For video generation training conditioned on multiple modalities, we include reference images, first frames, audio, and text as modal conditions. A multiple conditions dropout strategy is applied during training to enhance robustness. Specifically, text and audio are dropped with a 10\% probability independently. Meanwhile, the reference image and first frame are each dropped with a 20\% probability.  

To train pose-guided refiner, we also use Goku-I2V as pretrained base model. We adopt the training strategy from Humandit~\cite{gan2025humandit}, exposing the model to a range of resolutions to enable effective learning across diverse video qualities and sizes. Our conditioning modalities include pose extracted via Sapines~\cite{khirodkar2024sapiens}, first-frame reference images, and low-resolution 3D VAE latents. During training, a dropout mechanism is applied: both pose and low-resolution latents are dropped with a 20\% probability.

Both two models are trained using 128 NVIDIA GPUs with a learning rate of 5e-5. For LR-A2V inference, the we apply audio and text classifier-free guidance (CFG)~\cite{ho2022classifier} set to 6.5 and 30 denoising steps. For PG-Refiner, we apply pose CFG set to 1.5 and 20 denoising steps. Furthermore, we distill the PG-Refiner into a 1-step model while preserving output quality, enabling ultra-fast low-resolution generation and efficient high-resolution refinement with minimal steps. Detailed inference speed comparisons are provided in the appendix.

\vspace{-2mm}
\subsection{Comparison with State-of-the-Art Methods}
\label{sec:sota_comparison}

\noindent\textbf{Evaluation Metrics.}
%%short
To evaluate our model, we use a comprehensive video quality metric combining FID~\cite{heusel2017gans} for image quality, FVD~\cite{unterthiner2018towards} for video dynamics, and Q-align~\cite{wu2023q} for visual quality (IQA) and aesthetic appeal (AES). Lip-sync accuracy is assessed using Sync-C and Sync-D~\cite{chung2017out}, while identity consistency is measured with FaceSIM~\cite{yuan2025identity,huang2024consistentid}. For hand evaluation, we use average Hand Keypoint Confidence (HKC) and Hand Keypoint Variance (HKV).

\noindent\textbf{Test Datasets \& Baselines.}
For evaluation, we use the EMTD~\cite{echomimicv2} dataset, which contains 110 720P speech videos covering the upper body and hands. The longest video lasts 74 seconds, with 23.64\% of the videos exceeding 15 seconds, making it well-suited for assessing audio-driven portrait video generation in high-resolution, long-duration scenarios. To further evaluate the generalization ability of our method, we additionally select 100 samples from the HDTF~\cite{zhang2021flow} dataset at a resolution of 512$\times$512 as a talking-face test set. We also conduct a user test, detailed in the appendix.

We compare InfinityHuman with human animation methods, including FantasyTalking~\cite{wang2025fantasytalking}, Hallo3~\cite{hallo3}, HunyuanAvatar~\cite{chen2025hunyuanvideoavatar}, MultiTalk~\cite{kong2025let}, and OmniAvatar~\cite{gan2025omniavatar}, evaluated on the EMTD dataset. Since OmniHuman~\cite{lin2025omnihuman} is limited to 15-second videos and lacks long-form continuation support, it is excluded from the long-video evaluation. For short-video comparison, please refer to the appendix. In addition, we evaluate our method on the talking face generation benchmark HDTF~\cite{zhang2021flow}, comparing it with methods such as SadTalker~\cite{zhang2023sadtalker}, Aniportrait~\cite{wei2024aniportrait}, V-express~\cite{wang2024v}, EchoMimic~\cite{chen2025echomimic}, and other representative full-body models.

\input{table/ablation}
\noindent\textbf{Qualitative Results \& Quantitative Results.}
% For quantitative comparison, as shown in Table~\ref{tab:baseline}, our method achieves the best results in both FID and FVD for the audio-driven head and full-body animation tasks, demonstrating that our model produces more realistic and temporally coherent videos. Notably, on the EMTD dataset for full-body animation, our method significantly outperforms other approaches in terms of ID consistency (measured by FSIM) and hand quality (measured by the HKC). This confirms the effectiveness of the pose-guided refiner in maintaining consistent identities and highlights the advantages of incorporating hand-specific reward feedback learning.
For quantitative comparison, as shown in Table~\ref{tab:baseline}, our method achieves the best results in both FID and FVD across the audio-driven head and full-body animation tasks. Specifically, on the EMTD dataset, our model achieves an FID of 60.71 and an FVD of 979.88, outperforming the previous best results of 82.54 (OmniAvatar) and 1040.43 (MultiTalk), respectively. Notably, in full-body animation, our method achieves stronger identity consistency, with a FaceSIM of 0.84 (vs. 0.73 for Hallo3). It also delivers better hand motion quality, reaching the highest HKC of 0.90. These improvements demonstrate that our model generates videos that are both more visually realistic and exhibit better temporal coherence.

Additionally, we conduct a qualitative evaluation, as illustrated in Figure~\ref{fig:baseline_sota}. Our method demonstrates the ability to generate highly consistent and visually coherent animations over long sequences, maintaining a strong alignment with the corresponding audio. For instance, in the 40-second case, our approach ensures consistent id preservation and color harmony throughout the video. In contrast, other methods exhibit noticeable discrepancies in skin tone, hair color, and facial shapes, especially during long video continuations.

Our method also excels in hand generation, particularly when handling complex hand movements. While other models often struggle with severe distortions or unnatural gestures, our method ensures stable and realistic hand movements, even in challenging poses like hand crossing. This further underscores the superiority of our approach in managing intricate visual dynamics.

\input{figures/ablation}
\subsection{Ablation Study And Discussion}
\label{subsec:ablation}
\noindent\textbf{Ablation on the pose-guided refiner.}
By removing the pose-guided refiner, we directly decode videos from the low-resolution generator on a subset of the EMTD dataset to evaluate its effectiveness. As shown in Table~\ref{tab:ablation}, the overall generation quality significantly degrades, with FID increasing from 91.74 to 109.54 and FSIM dropping from 0.88 to 0.79. As illustrated in Figure~\ref{fig:ablation}, the degradation is particularly evident in blurred facial details and reduced temporal consistency. These results highlight the critical role of the refiner in recovering visual quality, enhancing temporal stability, and preserving identity over long sequences.

Furthermore, given that the refiner relies on multiple conditional inputs with non-trivial interdependencies, we conduct a deeper analysis of their individual contributions and guidance strength. As shown in Figure~\ref{fig:ablation}, omitting either the pose information or low-resolution latent features after training leads to color shifts and structural degradation in long-term video generation. This suggests that both inputs serve as essential references: the pose offers accurate structural constraints, while the low-resolution latent helps preserve overall semantic content and stylistic consistency.

\noindent\textbf{Ablation on the hand-specific reward feedback learning.}
We also assess the impact of the hand-specific reward feedback (refl) mechanism on generation performance. As shown in Table~\ref{tab:ablation}, removing it from the full model results in a decline in hand keypoint accuracy, with HKC decreasing from 0.87 to 0.85. Qualitatively, more artifacts and discontinuities appear in the hand regions, especially in sequences involving complex or high-speed gestures. These findings demonstrate that the hand-specific reward plays a vital role in improving the realism, stability, and audio synchronization of hand motion, particularly under challenging gestural conditions.

%% file: table/ablation.tex
\begin{table}[t]
\small
\centering
% \caption{\textbf{Quantitative Ablation Study.} Demonstrating the effectiveness of the pose-guided refiner and its corresponding conditions, including low-resolution video latent condition, pose guidance condition (\S\ref{subsec:pose_guided_refiner}), and hand-specific refl (\S\ref{subsec:hand_correction}).}
\vspace{-2mm}
% \label{tab:ablation}
\resizebox{\linewidth}{!}{
\begin{tabular}{|l|c|c|c|c|}
\Xhline{1.2pt}
\rowcolor{mygray}
Method & FID$\downarrow$ & FVD$\downarrow$ & FSIM$\uparrow$ & HKC$\uparrow$ \\
\hline
w/o refiner        & 109.54 & 876.49 & 0.79 & 0.85 \\
w/o lr cond       & 91.92 & 1001.00 & 0.86 & 0.85 \\
w/o pose cond    & 156.74 & 1163.75 & 0.83 & 0.83\\
w/o hand refl     & 86.32 & 844.57 & 0.86 & 0.85 \\
\rowcolor{mygray1}
ours               & \textbf{91.74} & \textbf{758.98} & \textbf{0.88} & \textbf{0.87} \\
\hline
\end{tabular}
}
\caption{\textbf{Quantitative Ablation Study.} Demonstrating the effectiveness of the pose-guided refiner and its corresponding conditions, including low-resolution video latent condition, pose guidance condition (\S\ref{subsec:pose_guided_refiner}), and hand-specific refl (\S\ref{subsec:hand_correction}).}
\vspace{-4mm}
\label{tab:ablation}
\end{table}

%% file: figures/ablation.tex
\begin{figure}[h]
\vspace{-6pt}
    \centering
    \includegraphics[width=0.95\linewidth]{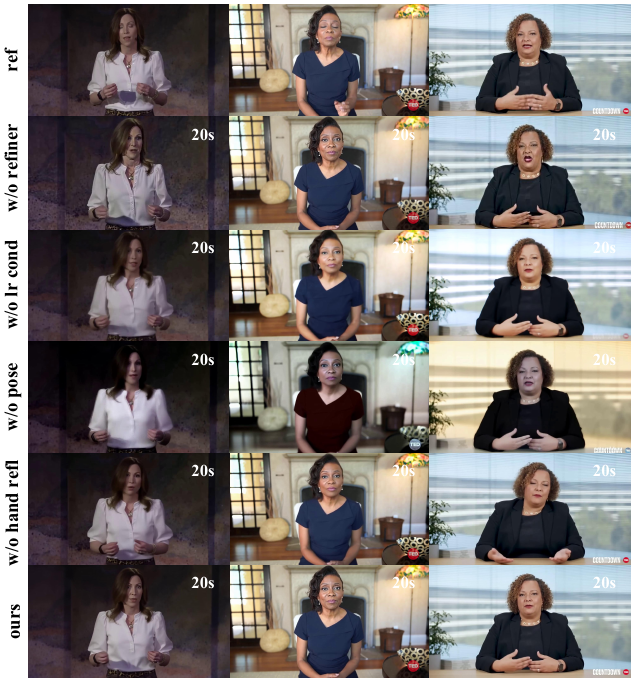}
\vspace{-2mm}
\caption{\textbf{Visualization of Ablation Study.} Demonstrating the effects of key components on animation quality.}
\label{fig:ablation}
\vspace{-4mm}
\end{figure}

%% file: sec/5_conclusion.tex
\vspace{-2mm}
\section{Conclusion and Future Work}
We present InfinityHuman, a coarse-to-fine framework for high-fidelity, long-duration, audio-driven full-body human animation. By introducing a pose-guided refiner and a hand-specific reward mechanism, our approach effectively addresses key challenges in visual consistency, lip-sync accuracy, and hand motion realism. Extensive experiments on EMTD and HDTF demonstrate that InfinityHuman achieves state-of-the-art performance across multiple metrics. 

% A limitation of our current framework is that the pose-guided refiner is trained solely on single-person data, which restricts its ability to handle multi-person interactions or dialogue scenarios. Additionally, our method supports generation consistent with the given reference but does not yet handle complex scene changes such as shot transitions or camera cuts. Extending InfinityHuman to support multi-person generation and dynamic scene transitions is an important direction for future work.

A limitation of our current framework is that it is trained solely on continuous single-person footage, which restricts its ability to handle multi-person interactions and complex scene transitions such as shot changes or cuts. Extending InfinityHuman to support multi-person generation and dynamic scene transitions is an important direction for future work.

%% file: sec/7_supp.tex
\newpage
\section{Appendix}
In this supplementary material, we provide additional technical details, experiment results, and design considerations that support the main paper. The content is organized as follows:
\begin{itemize}
    \item Short Baseline Comparisons (Sec.~\ref{sec:baseline});
    \item Detailed Parameters and Training Strategies (Sec.~\ref{sec:details});
    \item Multi-person Video Discussion and Solutions (Sec.~\ref{sec:multi});
    \item Speed Benchmark (Sec.~\ref{sec:speed});
    \item Hand-Specific Refl Generalization (Sec.~\ref{sec:reward});
    \item User Study Results (Sec.~\ref{sec:userstudy});
    \item Long-Form Video Stability (Sec.~\ref{sec:long});
    \item Societal Impacts (Sec.~\ref{sec:limitation}).
    \item More Demo Cases (Sec.~\ref{sec:demo});
\end{itemize}

%%%%%%%%%%%%%%%
\subsection{Short Baseline Comparisons}
\label{sec:baseline}

\begin{figure}[h]
    \centering
    \includegraphics[width=\linewidth]{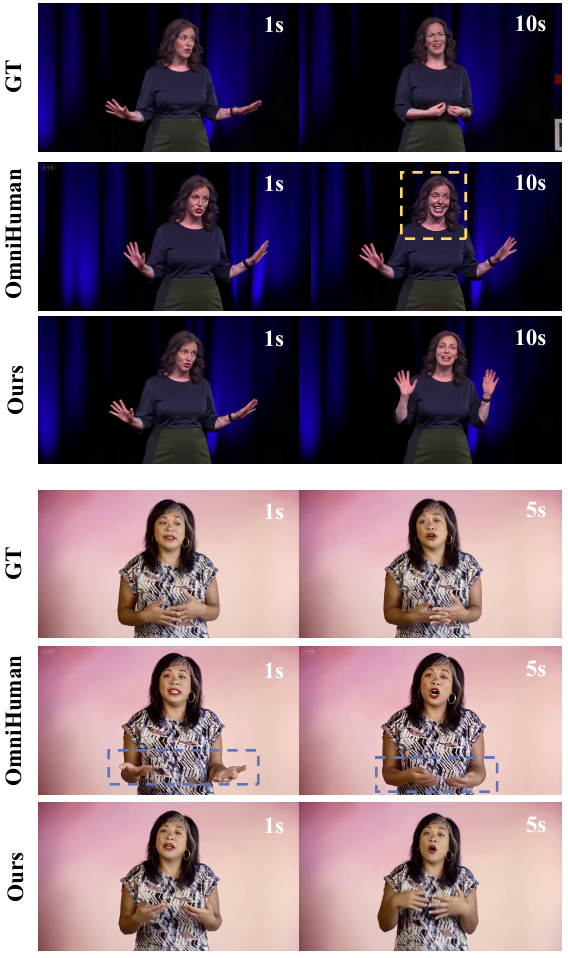}
    \caption{Qualitative comparison between OmniHuman and our method on 15s clips. Our model preserves finer facial details and produces more stable hand movements.}
    \label{fig:short_qualitative}
    \vspace{-4mm}
\end{figure}

OmniHuman~\cite{lin2025omnihuman} is limited to generating videos up to 15 seconds in length and lacks support for long-form temporal modeling, making it unsuitable for long-video evaluations presented in the main paper. Hence, we provide a dedicated comparison in the short-video setting (under 15 seconds).

\begin{table}[h]
    \centering
    % \caption{Quantitative comparison on short videos (under 15s) with OmniHuman.}
    % \label{tab:short_baseline}
    \resizebox{\linewidth}{!}{%
    \begin{tabular}{l|c|c|c|c|c}
        \toprule
        Model & FID $\downarrow$ & FVD $\downarrow$ & FSIM $\uparrow$ & HKC $\uparrow$ & Sync-C $\uparrow$ \\
        \midrule
        OmniHuman & 89.91 & 723.14 & 0.81 & 0.886 & 8.14 \\
        Ours & \textbf{77.05} & \textbf{540.89} & \textbf{0.88} & \textbf{0.894} & \textbf{8.98} \\
        \bottomrule
    \end{tabular}
    }
    \vspace{-2mm}
    \caption{Quantitative comparison on short videos (under 15s) with OmniHuman.}
    \label{tab:short_baseline}
    \vspace{-2mm}
\end{table}

As shown in Table~\ref{tab:short_baseline}, our model consistently outperforms OmniHuman across all evaluation metrics on short video generation. While OmniHuman achieves reasonable quality in short segments, our method delivers higher visual fidelity (lower FID and FVD), better structural similarity (FSIM), improved keypoint consistency (HKC), and stronger audio-visual synchronization (Sync-C), indicating enhanced short-term realism and coherence.

Figure~\ref{fig:short_qualitative} provides a qualitative frame-by-frame comparison. OmniHuman occasionally exhibits identity inconsistency and distorted hand shapes, whereas our method generates more temporally consistent gestures and better-preserved facial features throughout the sequence.

%%%%%%%%%%%%%%%
\subsection{Detailed Parameters and Training Strategies}
\label{sec:details}

% In our hand-specific mechanism, we utilize a curated hand dataset consisting of 10,000 samples, derived from a subset of our training data. For aesthetic evaluation, the T-score threshold was empirically set to 0.4. In addition, our dataset includes multilingual content, enabling the model to support input in various languages and generate corresponding results across different linguistic contexts.
% xiepan
For hand-specific reward model, to construct data pairs for training a "hand distortion" dimension, we follow a methodology inspired by the Multi-dimensional Human Preference (MHP) dataset~\cite{zhang2024learning} construction process, with specific adaptations for hand-related evaluation. First, prompt collection focuses on scenarios involving hands to ensure targeted data generation. Next, image generation utilizes several state-of-the-art text-to-image models. Each model generates 2–4 images per prompt, yielding approximately 40,000 images. This diversity ensures coverage of varying hand distortion patterns across model types and settings. For annotation, a crowdsourced team of 10 annotators is trained with clear guidelines: hand distortion is defined as abnormal finger count, twisting, blurriness, structural incoherence (e.g., missing joints), or misalignment with context (e.g., a "grasping" hand without closed fingers). Annotators rate each image in a pair on a 1–5 scale (1 = severe distortion, 5 = perfect integrity) for the "hand distortion" dimension.

For aesthetic evaluation, the T-score threshold was empirically set to 0.4. In addition, our dataset includes multilingual content, enabling the model to support input in various languages and generate corresponding results across different linguistic contexts.

In our Low-Resolution Noise Injection Strategy, Gaussian noise is added to the low-resolution encoder and scaled by a factor $\alpha$. We found that $\alpha = 0.7$ achieves the best trade-off between visual realism and temporal stability. Consistent with observations in SimpleGVR~\cite{xie2025simplegvr}, we noted that training with small noise levels ($\alpha \in [0.0, 0.3]$) led to suboptimal refinement of fine details. In contrast, using higher noise levels ($\alpha \in [0.6, 0.9]$) enabled the model to better recover detail structures, especially when guided by pose estimation.

For pose extraction, we employ the Sapiens estimator~\cite{khirodkar2024sapiens}, which demonstrates robust performance across a wide range of motion types.

% For evaluation, we used different text prompts across datasets. The prompt for HDTF is:  
% \textit{“A realistic video of a man/woman speaking directly to the camera. His/Her facial expressions are expressive and full of emotion, enhancing the delivery. The camera remains steady, capturing sharp, clear movements and a focused, engaging presence.”}

% The prompt for EMTD is more gesture-centric:  
% \textit{“A realistic video of a man/woman speaking directly to the camera, waving his/her hand with dynamic and rhythmic hand gestures that complement his/her speech. His/Her hands are clearly visible, independent and unobstructed. His/Her facial expressions are expressive and full of emotion, enhancing the delivery. The camera remains steady, capturing sharp, clear movements and a focused, engaging presence.”}

For evaluation, we used different prompts across datasets. The EMTD prompt is:  
\textit{“A realistic video of a man/woman speaking directly to the camera, waving his/her hand with dynamic and rhythmic hand gestures that complement the speech. His/Her hands are clearly visible, independent and unobstructed. Facial expressions are expressive and full of emotion, enhancing the delivery. The camera remains steady, capturing sharp, clear movements and a focused, engaging presence.”}  
For HDTF, we use the same prompt without the hand gesture sentence.

%%%%%%%%%%%%%%%
% \vspace{-3mm}
\subsection{Multi-person Video Discussion and Solutions}
% \vspace{-3mm}
\label{sec:multi}

\begin{figure}[h]
    \centering
    \includegraphics[width=0.6\linewidth]{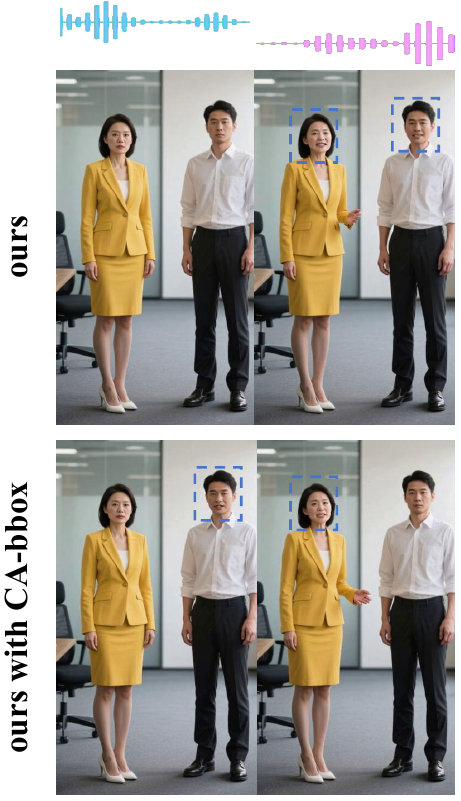}
    \caption{Illustration of multi-person alternating speaking. Without control, both characters speak simultaneously. With audio-aware attention and bounding box assignment, only the target character moves, while the other remains idle.}
    \vspace{-2mm}
    \label{fig:sup_multi}
    \vspace{-2mm}
\end{figure}

Handling multi-person scenarios introduces challenges such as accurate pose assignment, occlusion management, and audio-motion alignment. Since our training data primarily consists of single-person, continuous video clips, directly applying our model to reference images containing multiple individuals and alternating speech audio may result in incorrect behavior—such as both people speaking simultaneously. As shown in Fig.~\ref{fig:sup_multi}, when no control is applied, both characters erroneously respond to the same speech input.

To address this, we propose a simple extension based on fixed bounding boxes and an audio-aware attention mechanism to preliminarily support multi-person video generation. Specifically, for each human bounding box (e.g., bbox1 and bbox2), we introduce directional audio control in the cross-attention module. For the currently speaking character (e.g., bbox1), the visual features are attended to using the actual speech audio features; for the non-speaking character (e.g., bbox2), we input silent audio features to suppress unintended movements. This approach enables synchronized turn-taking behavior between characters without modifying the core network architecture. As shown in Fig.~\ref{fig:sup_multi}, our strategy enables clear alternation of speaking roles between multiple characters.

However, because our model is not trained on multi-person data, it lacks the ability to handle rich interactions such as mutual gaze, coordinated gestures, or conversational feedback. Additionally, the current Pose-Guided Refiner module, which relies on additive conditioning, is not yet capable of processing multiple pose streams simultaneously.

In summary, while our approach provides a lightweight and scalable extension for basic multi-person, alternating-speech scenarios, generating dynamic, interactive multi-character talking videos remains a challenging direction for future research. Progress in dataset construction and model architecture will be crucial to further advance this capability.

%%%%%%%%%%%%%%%
\vspace{-2mm}
\subsection{Speed Benchmark}
\label{sec:speed}

% xiepan
We applied distillation acceleration on 720p-Refiner. First, we employed PCM~\cite{wang2024phased,ren2024hyper} for step distillation, compressing it to 4 steps inference. Then, we used LADD~\cite{sauer2024fast} for adversarial training. Finally, we achieve Refiner-distill with one-step inference.

We selected a 720P-resolution subset from our validation set to benchmark the inference efficiency and generation quality of different models. Specifically, we compared our model before and after knowledge distillation, evaluating both runtime and quality metrics.

Table~\ref{tab:speed} summarizes the inference time and quality metrics (FID and FVD) across models on this subset. Our distilled model significantly reduces inference time while maintaining stable performance compared to the pre-distilled version.

This efficiency gain is attributed to our coarse-to-fine architecture: the majority of denoising steps are performed at a lower resolution (e.g., 360P), with only a few refinement steps conducted at the high-resolution stage. This design effectively reduces computation while preserving final output fidelity, demonstrating the practical advantages of our method in real-time generation scenarios.
\begin{table}[h]
    \centering
    % \caption{Inference Time and Quality Metrics. Distilled model reduces time while maintaining quality.}
    % \label{tab:speed}
    \resizebox{\linewidth}{!}{%
    \begin{tabular}{l|c|c|c|c}
        \toprule
        Model & Aud. (s) & Inf. Time (s) & FID $\downarrow$ & FVD $\downarrow$ \\
        \midrule
        LR-A2V + Refiner & 16.4 & 552 & 50.51 & 909.82 \\
        LR-A2V + Refiner-distill & 16.4 & 187 & 48.17 & 884.27\\
        \bottomrule
    \end{tabular}
    }
    \vspace{-2mm}
    \caption{Inference Time and Quality Metrics. Distilled model reduces time while maintaining quality.}
    \vspace{-2mm}
    \label{tab:speed}
    % \vspace{-2mm}
\end{table}

\begin{figure}[h]
    \centering
    \includegraphics[width=0.8\linewidth]{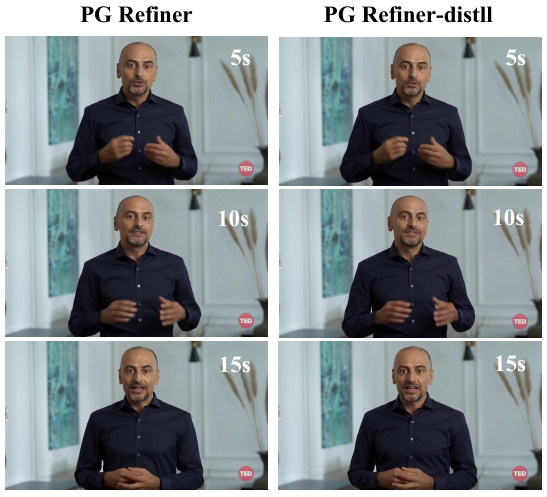} 
    % \vspace{-6mm}
    \caption{Visual comparison between the original and distilled model outputs. Perceptual difference is minimal, though the distilled version tends to produce slightly sharper foregrounds and less blended backgrounds. \textit{Please zoom in for detailed inspection.}}
    \label{fig:distill_vis}
    \vspace{-6mm}
\end{figure}

As shown in Figure~\ref{fig:distill_vis}, the perceptual quality between the distilled and original versions is nearly indistinguishable. While the distilled model may produce slightly sharper edges and less background blending in some frames, the overall visual fidelity remains high.

%%%%%%%%%%%%%%%
\vspace{-3mm}
\subsection{Hand-Specific Refl Generalization}
\label{sec:reward}

\begin{figure}[h]
    \centering
    \includegraphics[width=1.0\linewidth]{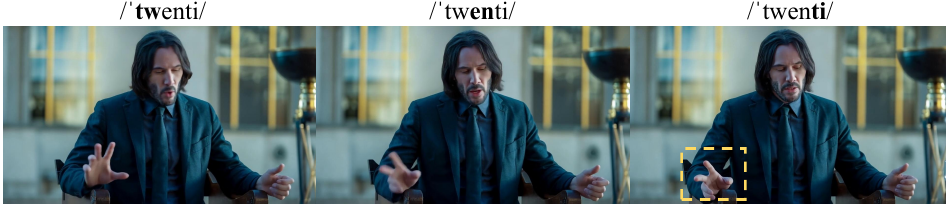}
    \caption{When the character says \textit{"twenty"}, the hand performs a two-finger gesture, demonstrating alignment between speech and hand gesture.}
    \vspace{-2mm}
    \label{fig:twenty_gesture}
    \vspace{-2mm}
\end{figure}

Our hand-specific feedback reward mechanism demonstrates strong generalization to unseen gestures. During testing, we observed that it maintains a high degree of alignment with the accompanying text. As illustrated in Figure~\ref{fig:twenty_gesture}, when the character in the video says the word \textit{"twenty"}, the hand simultaneously performs a gesture corresponding to the number two, showcasing a remarkable level of semantic and temporal coherence. Notably, when we remove the proposed hand-specific reward mechanism, such gestures no longer emerge, indicating that our method significantly enhances the correspondence between hand motion and spoken content.

%%%%%%%%%%%%%%%
\vspace{-2mm}
\subsection{User Study Results}
\label{sec:userstudy}

To evaluate the perceptual quality of the generated videos, six expert raters assessed 78 videos from two methods: InfinityHuman and Omnihuman~\cite{lin2025omnihuman}. The evaluation covered overall video quality (high-standard and non-high-standard), compliance (e.g., stutter, clarity, commonsense), facial naturalness (hands, mouth, eyes), background distortion, and visual consistency (color, stability). Ratings were categorized as Excellent, Qualified, or Unqualified. The summarized results are presented in Table~\ref{tab:userstudy_results}.
\begin{table}[h]
\centering
% \caption{User study results: counts and percentages (\%) of rated videos.}
% \label{tab:userstudy_results}
\resizebox{\linewidth}{!}{%
\begin{tabular}{l|c|c}
\toprule
Category & Ours & OmniHuman \\
\midrule
Total Videos & 78 (100\%) & 78 (100\%) \\
High-Standard Qualified & 13 (16.7\%) & 0 (0.0\%) \\
High-Standard Unqualified & 65 (83.3\%) & 78 (100\%) \\
Non-High-Standard Qualified & 18 (23.1\%) & 0 (0.0\%) \\
Non-High-Standard Unqualified & 60 (76.9\%) & 78 (100\%) \\
Video Compliance Qualified & 78 (100\%) & 78 (100\%) \\
Video Clarity Qualified & 78 (100\%) & 77 (98.7\%) \\
Hand Naturalness Unqualified & 2 (2.6\%) & 41 (52.6\%) \\
Mouth Naturalness (Excellent) & 9 (11.5\%) & 7 (9.0\%) \\
Background Distortion Unqualified & 4 (5.1\%) & 18 (23.1\%) \\
Image Stability Unqualified & 1 (1.3\%) & 64 (82.1\%) \\
\bottomrule
\end{tabular}
}
\vspace{-2mm}
\caption{User study results: counts and percentages (\%) of rated videos.}
\label{tab:userstudy_results}
\vspace{-4mm}
\end{table}

Overall, our method shows clear advantages in hand naturalness and image stability, with low failure rates of 2.6\% and 1.3\%, respectively, compared to significantly higher failure rates for Omnihuman. Besides, our method achieves a more balanced overall quality and background consistency, highlighting its strengths in producing stable and natural outputs.

%%%%%%%%%%%%%%%
\vspace{-2mm}
\subsection{Long-Form Video Stability}
\label{sec:long}

To explore the stability of our model on long-form video generation, we segment each output into consecutive 10-second clips and compute cumulative metrics over time (i.e., 10s, 20s, 30s, etc.). This progressive evaluation enables us to analyze how performance evolves as video duration increases.

As shown in Table~\ref{tab:longform_metrics}, our model maintains stable performance throughout extended video lengths. Specifically, key metrics such as FID, FVD, FSIM, and Sync show minimal degradation, indicating strong temporal consistency and robustness. In contrast, baseline models tend to suffer from more noticeable quality drops as duration increases.

\begin{table}[h]
    \centering
    % \caption{Cumulative evaluation on long-form video generation over increasing durations.}
    % \label{tab:longform_metrics}
    \resizebox{\linewidth}{!}{%
    \begin{tabular}{l|c|c|c|c|c}
        \toprule
        Duration & FID $\downarrow$ & FVD $\downarrow$ & FSIM $\uparrow$ & HKC $\uparrow$ & Sync-C $\uparrow$ \\
        \midrule
        10s & 36.83 & 1015.36 & 0.8357 & 0.9224 & 7.23 \\
        20s & 37.07 & 1156.05 & 0.8323 & 0.9062 & 7.36 \\
        30s & 35.02 & 1315.40 & 0.8266 & 0.8991 & 7.62 \\
        40s & 35.92 & 1260.71 & 0.8154 & 0.9007 & 7.81 \\
        50s & 35.50 & 945.84  & 0.8057 & 0.9059 & 7.46 \\
        \bottomrule
    \end{tabular}
    }
    \vspace{-2mm}
     \caption{Cumulative evaluation on long-form video generation over increasing durations.}
    \label{tab:longform_metrics}
    \vspace{-2mm}
\end{table}

%%%%%%%%%%%%%%%
\vspace{-2mm}
\subsection{Societal Impacts}
\label{sec:limitation}
Our method enables the generation of high-fidelity talking head videos with synchronized hand gestures, which can benefit applications such as digital avatars, virtual presenters, and language learning tools. However, the ability to synthesize realistic human figures also poses potential risks. Inappropriate or malicious use of such technology—such as creating misleading or non-consensual content—could have negative societal consequences. To mitigate misuse, we advocate for strong watermarking, provenance tracking, and ethical deployment practices. Moreover, datasets used for training should be carefully curated to avoid reinforcing harmful biases or stereotypes. Future work should continue to emphasize fairness, transparency, and responsible AI development.

%%%%%%%%%%%%%%%
\vspace{-2mm}
\subsection{More Demo Cases}
\label{sec:demo}
To further demonstrate the robustness and generality of our method, we provide several additional qualitative results across diverse scenarios.

\begin{figure*}[t]
    \centering
    \includegraphics[width=0.82\textwidth]{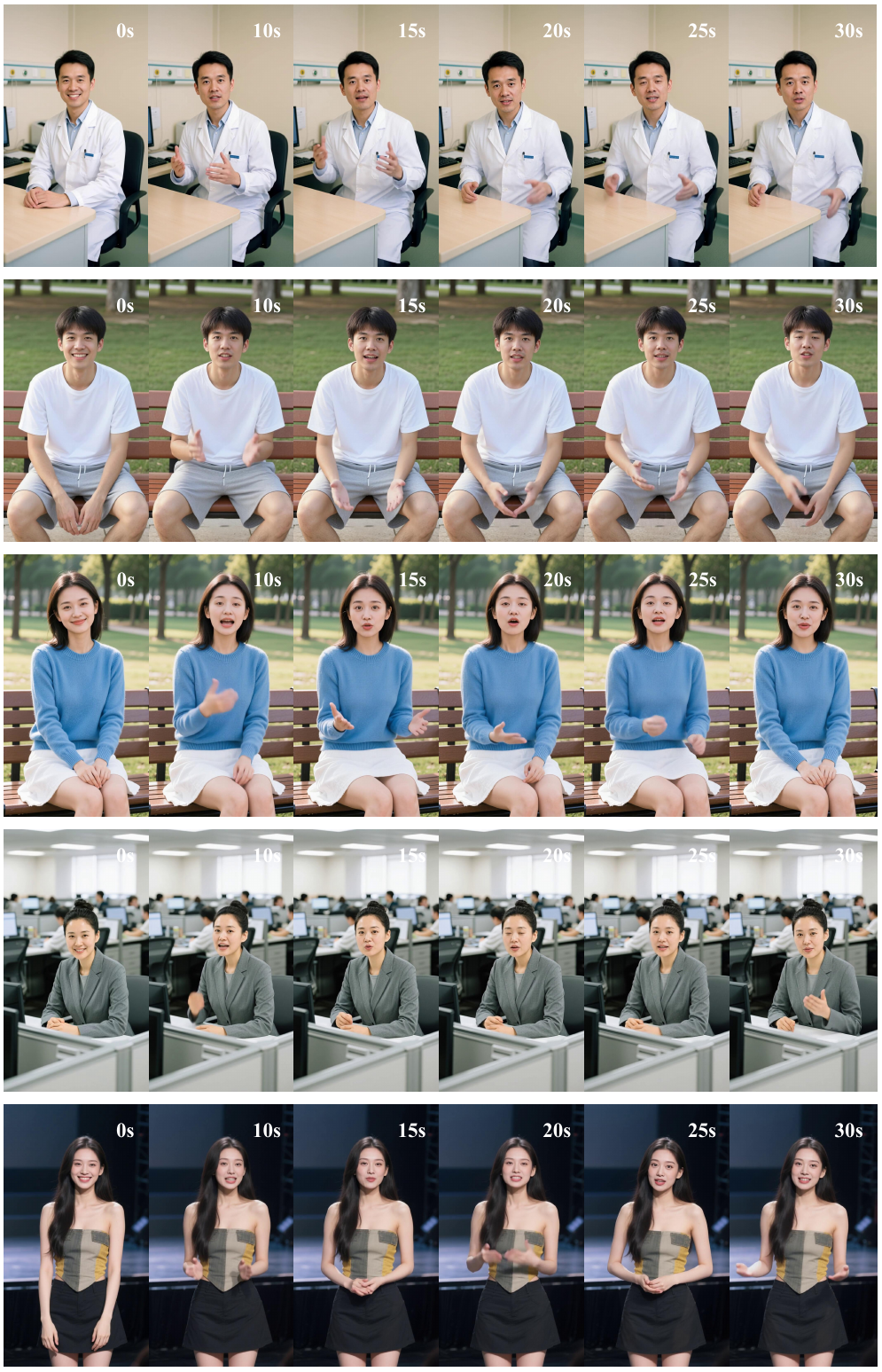} 
    \caption{Additional qualitative results showcasing the robustness and generality of our method across diverse scenarios.}
    \label{fig:generalization}
\end{figure*}